\documentclass[conference]{IEEEtran}
\IEEEoverridecommandlockouts
\usepackage{cite}
\usepackage{amsmath,amssymb,amsfonts}
\usepackage{algorithmic}
\usepackage{graphicx}
\usepackage{textcomp}
\usepackage{xcolor}
\usepackage{tabularx}
\usepackage{array}

\def\BibTeX{{\rm B\kern-.05em{\sc i\kern-.025em b}\kern-.08em
    T\kern-.1667em\lower.7ex\hbox{E}\kern-.125emX}}

\makeatletter
\newcommand{\newlineauthors}{%
\end{@IEEEauthorhalign}\hfill\mbox{}\par
\mbox{}\hfill\begin{@IEEEauthorhalign}
}
\makeatother

\begin{document}

\title{A Deep Learning Based Automated Hand Hygiene Training System\\
\thanks{Authors would like to acknowledge the financial support of Tavan Ressan Company.}
}

\author{
	\IEEEauthorblockN{Mobina Shahbandeh}
	\IEEEauthorblockA{\textit{School of Electrical and } \\ \textit{Computer Engineering} \\
		\textit{University of Tehran}\\
		Tehran, Iran \\
		mobinashahbandeh@ut.ac.ir}
	\and
	\IEEEauthorblockN{Fatemeh Ghaffarpour}
	\IEEEauthorblockA{\textit{School of Electrical and } \\ \textit{Computer Engineering} \\
		\textit{University of Tehran}\\
		Tehran, Iran \\
		f.ghaffarpour@ut.ac.ir}
	\and
	\IEEEauthorblockN{Sina Vali}
	\IEEEauthorblockA{\textit{Human and Robot} \\ \textit{Interaction Laboratory}\\
		\textit{University of Tehran}\\
		Tehran, Iran \\
		s.vali@mail.sbu.ac.ir}
	\and
	\IEEEauthorblockN{Mohammad Amin Haghpanah}
	\IEEEauthorblockA{\textit{School of Electrical and } \\ \textit{Computer Engineering} \\
		\textit{University of Tehran}\\
		Tehran, Iran \\
		amin.haghpanah@ut.ac.ir}
	\newlineauthors
	\IEEEauthorblockN{Amin Mousavi Torkamani}
	\IEEEauthorblockA{\textit{Department of} \\ \textit{Computational Engineering} \\
		\textit{Lappeenranta University of Technology}\\
		Lappeenranta, Finland \\
		amin.mousavi.torkamani@student.lut.fi}
	\and
	\IEEEauthorblockN{Mehdi Tale Masouleh}
	\IEEEauthorblockA{\textit{School of Electrical and } \\ \textit{Computer Engineering} \\
		\textit{Human and Robot Interaction} \\ \textit{Laboratory}\\
		\textit{University of Tehran}\\
		Tehran, Iran \\
		m.t.masouleh@ut.ac.ir}
	\and
	\IEEEauthorblockN{Ahmad Kalhor}
	\IEEEauthorblockA{\textit{School of Electrical and } \\ \textit{Computer Engineering} \\
		\textit{University of Tehran}\\
		Tehran, Iran \\
		akalhor@ut.ac.ir}
}

\maketitle

\begin{abstract}
	Hand hygiene is crucial for preventing viruses and infections. Due to the pervasive outbreak of COVID-19, wearing a mask and hand hygiene appear to be the most effective ways for the public to curb the spread of these viruses. The World Health Organization (WHO) recommends a guideline for alcohol-based hand rub in eight steps to ensure that all surfaces of hands are entirely clean. As these steps involve complex gestures, human assessment of them lacks enough accuracy. However, Deep Neural Network (DNN) and machine vision have made it possible to accurately evaluate hand rubbing quality for the purposes of training and feedback. In this paper, an automated deep learning based hand rub assessment system with real-time feedback is presented. The system evaluates the compliance with the 8-step guideline using a DNN architecture trained on a dataset of videos collected from volunteers with various skin tones and hand characteristics following the hand rubbing guideline. Various DNN architectures were tested, and an Inception-ResNet model led to the best results with 97\% test accuracy. In the proposed system, an NVIDIA Jetson AGX Xavier embedded board runs the software. The efficacy of the system is evaluated in a concrete situation of being used by various users, and challenging steps are identified. In this experiment, the average time taken by the hand rubbing steps among volunteers is 27.2 seconds, which conforms to the WHO guidelines.
\end{abstract}


\section{Introduction}
Contaminated hands of people, especially healthcare professionals, are the most common vehicle for the transmission of healthcare-associated pathogens. Hand hygiene has been proven to be effective in preventing healthcare-associated infections and diseases \cite{b1, b2, b3}, such as COVID-19 \cite{b4}, which is now widespread in the world, and preventing it has utmost importance. In this regard, alcohol-based hand rub has been demonstrated to have advantages over hand washing with soap \cite{b5}, and the World Health Organization (WHO) has provided a hand rub guideline in eight steps \cite{b6}. Compliance with this guideline should be monitored to ensure hand cleanliness. As human observation is not accurate enough, automated quality control of hand rub is mandatory to provide feedback and training. For this purpose, some existing research have utilized classic computer vision approaches. In \cite{b7}, a commercial product is built by Surewash using these traditional approaches for hand hygiene training, and the paper provides data about the effectiveness of providing user feedback on improving the medical staff's adherence to the WHO guideline in a hand washing task. On the other hand, Convolutional Neural Network (CNN) architectures are being used in a broad range of applications, including image classification. Therefore, CNNs can also be exploited in detecting hand hygiene steps.  In \cite{b8}, hand hygiene dispenser usage is detected using a CNN and depth sensor data. In \cite{b9}, hand washing compliance detection is also done using depth sensor data and a CNN. In \cite{b10}, recognizing a non-detail-level hand washing task's actions is performed using two CNNs. In \cite{b11} and \cite{b12}, compliance of hand washings with the WHO guidelines is assessed using data collected from a wearable sensor and Machine Learning techniques. Most existing research propounded in the literature leverage sensor data or classic machine vision techniques. In this paper, a state-of-the-art CNN architecture is employed for the real-time classification of user actions using camera images.

Detecting hand rub poses can be regarded as a gesture recognition task. Gesture recognition has been applied in various contexts. In \cite{b13}, a CNN is used to detect the American Sign Language gestures. In \cite{b14}, the rudimentary task of detecting the presence of one hand, two hands, and no hand is carried out using a pre-trained neural network. 

\begin{figure}[tb]
	\centerline{\includegraphics[width=\linewidth]{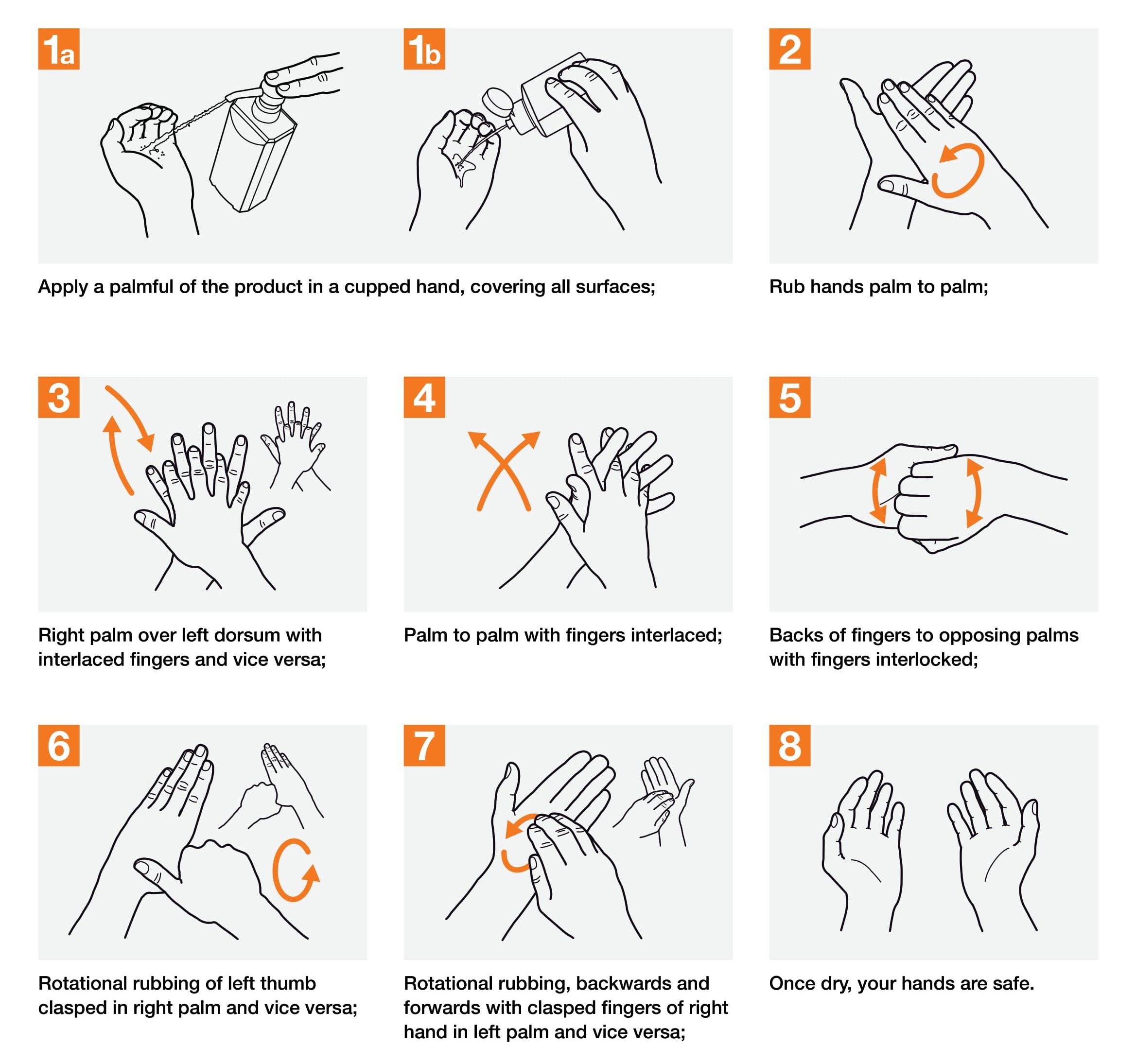}}
	\caption{WHO Guidelines on hand rub \cite{b6}, and in this paper steps 2-8 are taken into account.}
	\label{fig:guideline}
\end{figure}

In this work, an automated system called DeepHARTS (short for Deep learning based HAnd Rub Training System) is presented, which evaluates the conformance to the hand rubbing guideline by users with real-time feedback. The markerless gesture recognition of hand rubbing steps is carried out using a CNN architecture. The CNN is built on top of Tensorflow and is trained using a dataset containing videos of each hand rub step performed by 22 volunteers following the guideline provided by WHO. In the fabricated system, a camera is placed on the top of the surface where the user washes her hands. The frames of user activities are captured by the camera and then fed to the CNN. Then, a frame-based classification of the user action is performed by the network. The system guides the user through hand rub steps and informs the user whether the steps have been taken correctly or not using visual indicators on the screen. The system's efficiency is evaluated using data gathered from a real-life scenario. The challenges in making such a system and difficult hand rub poses are identified in this paper.

This paper includes the following contributions:

\begin{itemize}
	\item A deep learning based system for guiding users through the whole hand rubbing process using real-time feedback and interactive Graphical User Interface (GUI);
	\item Providing a dataset including videos of various volunteers following the WHO guideline on hand rub;
	\item An evaluation on real-world usage to assess the perceived efficacy of the proposed system;
	\item Identifying challenging steps in hand rubbing using data gathered from the application of the proposed system in a real-life scenario.
\end{itemize}

The remainder of this paper is organized as follows. Section II provides the hardware and software description of the system, including the details of the CNN model. The evaluation of the system is discussed in Section III. Finally, Section IV includes a conclusion with a summary and future work of this research.

\begin{figure}[t]
	\centerline{\includegraphics[height=8cm]{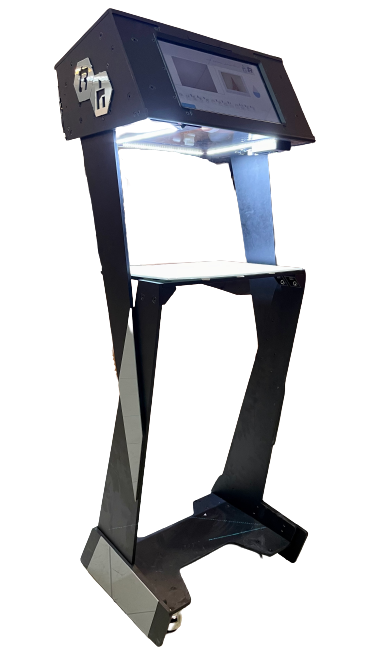}}
	\caption{A schematic of DeepHARTS.}
	\label{fig:system}
\end{figure}

\section{DeepHARTS System Description}

This section discusses the structure of DeepHARTS, including its mechanical, electrical, and software components.

\subsection{Mechanical Description}
\begin{figure}[t]
	\centerline{\includegraphics[width=5.9cm]{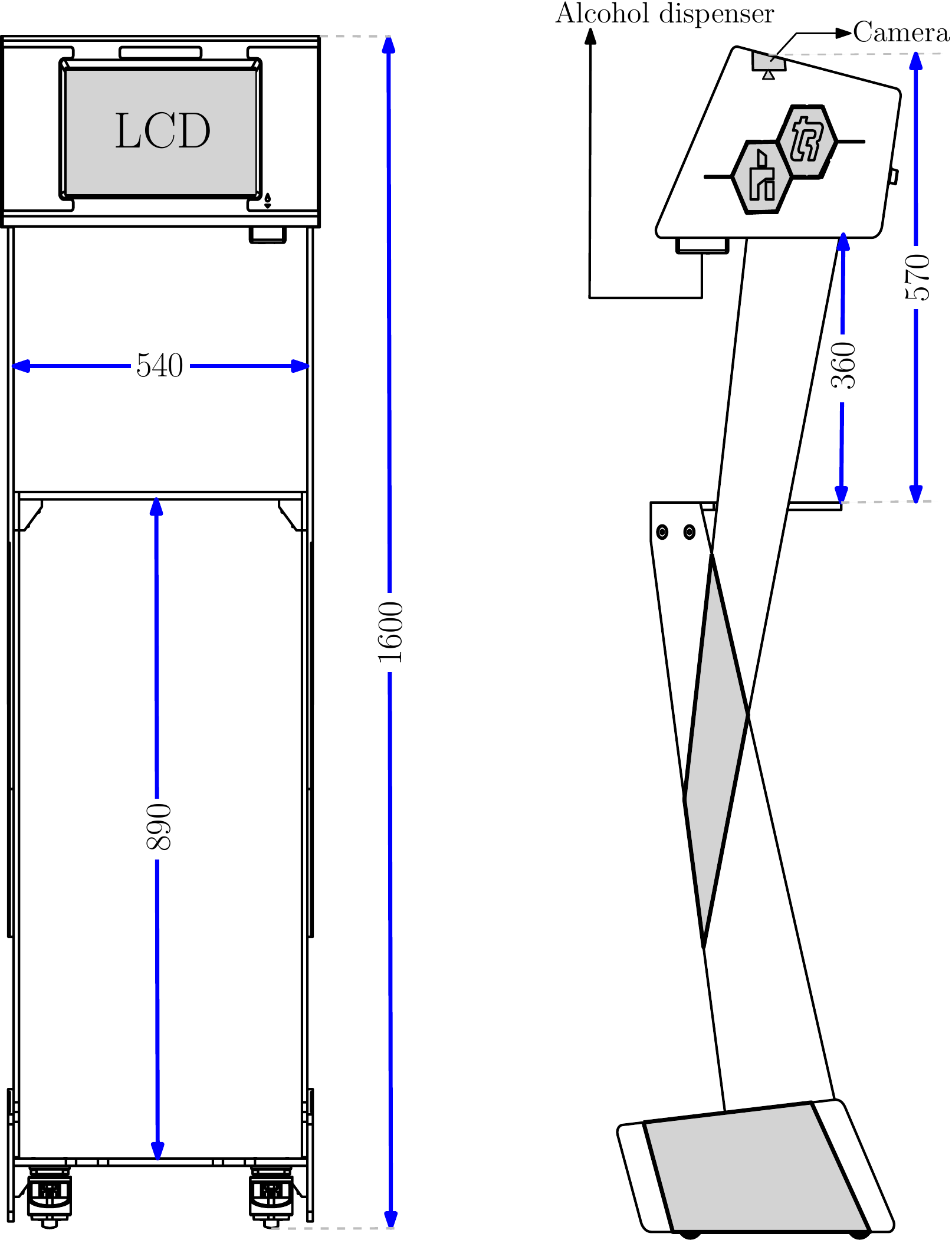}}
	\caption{DeepHARTS, all dimensions are in milimeters.}
	\label{fig:dims}
\end{figure}

Fig. \ref{fig:system} depicts the so-called DeepHARTS, which is an automated hand hygiene system developed by the Human and Robot Interaction Laboratory, University of Tehran, under a joint project with Tavan Ressan Company. The structure is made of steel, and its dimensions are illustrated in Fig. \ref{fig:dims}. It is equipped with four wheels for portability and a 15.5-inch touch screen for user convenience. From an ergonomic standpoint, prior to designing the structure, to find an appropriate dimension for the structure, a structure using aluminum profiles was fabricated, and several tests were performed. The accessibility to all system components is through the back, and filling the alcohol tank can be done from the top of the structure. The background of the video stream of the user’s hands is homogenous and white, which alleviates challenges involved in machine vision procedures, such as hand segmentation. By the same token, a light-emitting diode (LED) illuminates the background, leading to a uniform and unvaried color histogram for captured frames.

\begin{figure}[b]
	\centerline{\includegraphics[width=\linewidth]{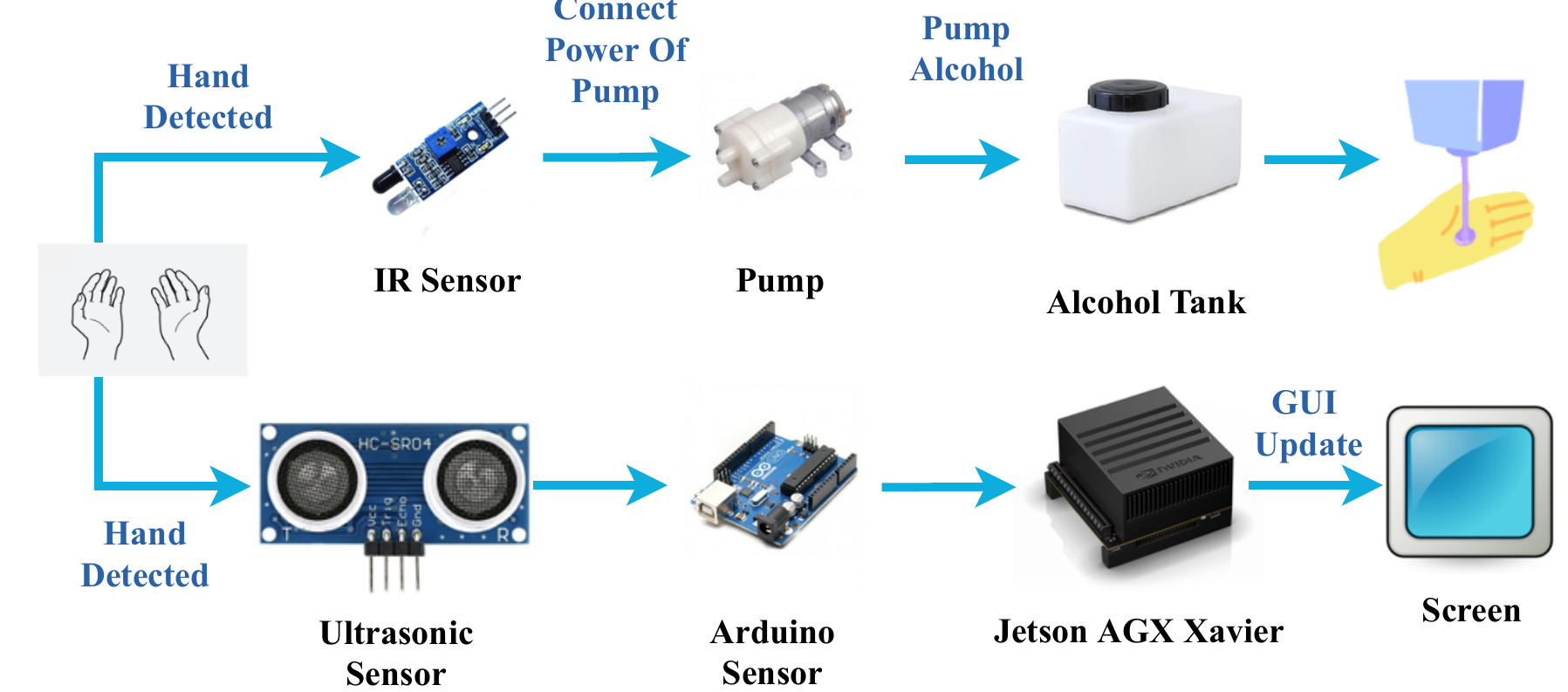}}
	\caption{The procedure of getting the sanitizer and providing a visual feedback.}
	\label{fig:sanitizerprocess}
\end{figure}

\subsection{Electrical Description}
A Deep Neural Network (DNN) is used in this system, which requires computation power and thus appropriate computer hardware. NVIDIA provides a wide variety of Artificial Intelligence computers. For the purposes of this project, two of these computers were available and tested. At an earlier stage, an NVIDIA Jetson Nano (4GB Memory) was used to run the DNN model for detecting the hand rubbing process. However, as the Jetson Nano lacks enough computation power to run such a deep network alongside the GUI, another board such as Raspberry Pi was required to run the GUI which will be discussed in Subsection C. Hence, the computer has now been replaced by an NVIDIA Jetson AGX Xavier (512-core Volta GPU, 8-core CPU, 32GB Memory).

\begin{figure}[t]
	\centerline{\includegraphics[width=\linewidth]{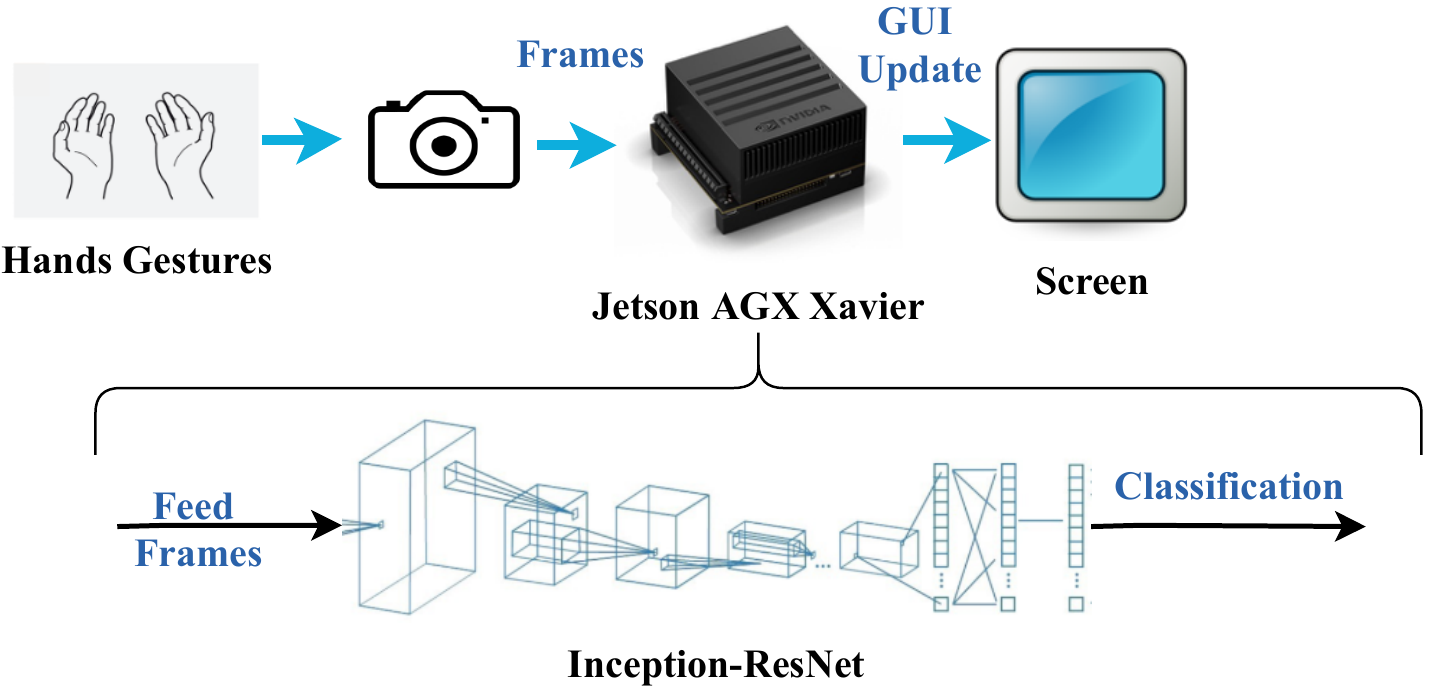}}
	\caption{Steps of the classification of user hands gesture and generating a visual feedback.}
	\label{fig:classificationprocess}
\end{figure}

The main procedures for detecting hand rubbing steps fall into three steps as follows.

\paragraph{Hand detection step} First, the user initiates a hand rubbing process by bringing her hands under the camera. The camera points down to the surface where the user washes her hands, and it captures the hand pose and passes the frames to the Jetson AGX Xavier. The user should place her hands in the region shown on the screen. The correctness of this step is detected by hand segmentation from the background using Open-Source Computer Vision (OpenCV), a library used for image processing. 

\paragraph{Sanitizer step} Fig. \ref{fig:sanitizerprocess} explains the procedure of getting the sanitizer and updating the GUI. A board is used for an automatic alcohol dispenser consisting of an IR sensor to sense the user’s hands. The system asks the user to bring her hands under the alcohol dispenser. When an IR sensor detects the user’s hands, it connects power to pump alcohol from the alcohol tank. Simultaneously, an Ultrasonic sensor next to the alcohol dispenser measures the distance from the object to which the sensor is faced. This sensor is connected to an Arduino ATmega board, which is programmed to calculate the distance of an object and constantly report the computed distance to the Jetson AGX Xavier. If the measured distance is in a defined range, the system detects that the object is the user’s hands placed under the sensor. Thus, the screen is updated, and the step of getting the sanitizer is passed.

\paragraph{Hand rubbing steps} After finishing the two aforementioned steps, the system guides the user through hand rubbing steps. The user starts hand rubbing based on steps 2 to 7 of the guideline as depicted in Fig. \ref{fig:guideline}. Also, Fig. \ref{fig:classificationprocess} shows the steps of classifying user hands gesture. First, the camera captures several frames from the hand pose and passes the frames to the Jetson AGX Xavier. Then, the frames get fed to the CNN that performs the classification. If the predicted step has a high probability, the step is considered passed, and the GUI shows corresponding feedback to the user. Otherwise, the user should repeat the step. After every three steps as illustrated in Fig. \ref{fig:guideline}, the system asks the user to repeat the sanitizer step.

\begin{figure}[t]
	\centerline{\includegraphics[width=\linewidth]{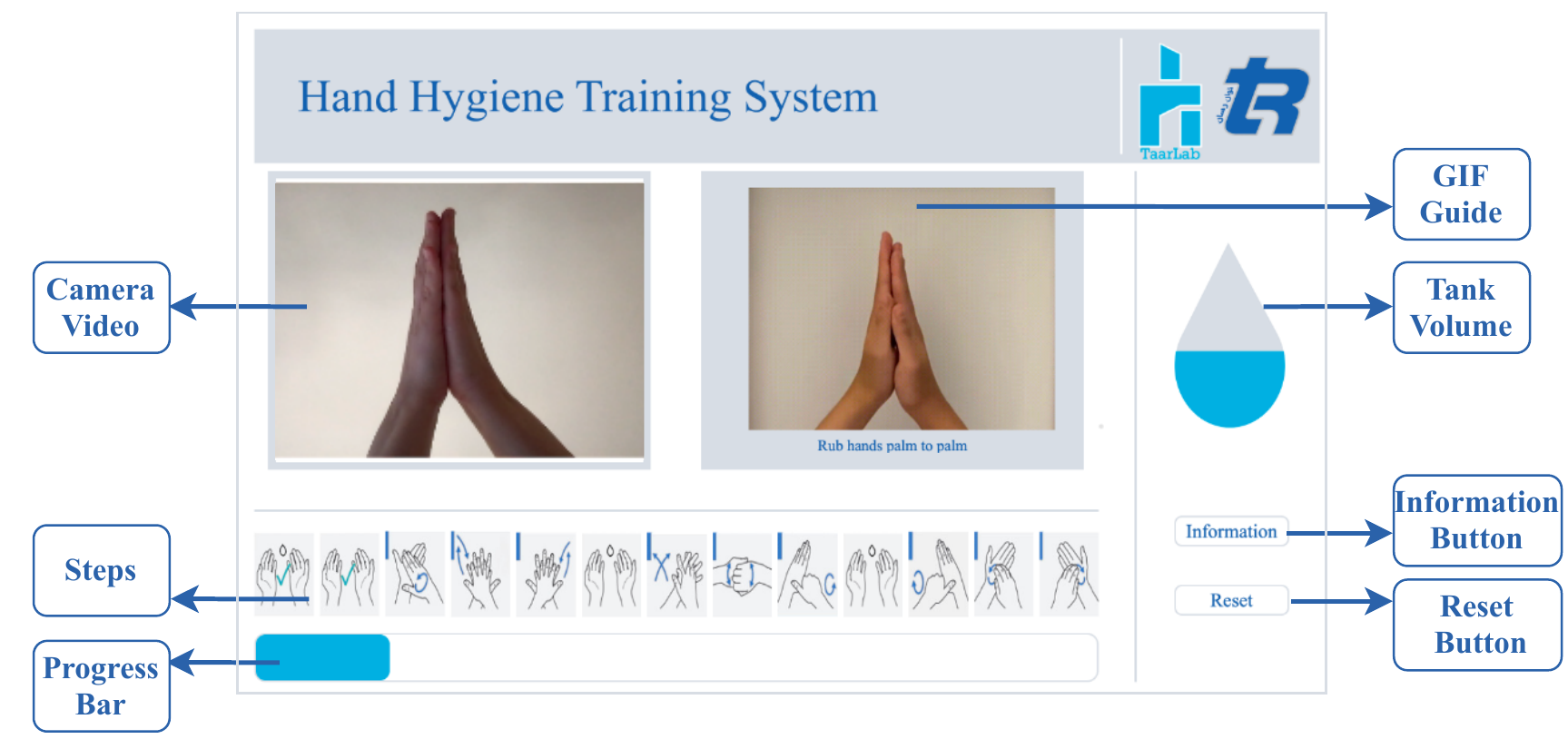}}
	\caption{The main screen of the user interface.}
	\label{fig:gui}
\end{figure}

\begin{table*}[t]
	\caption{Performance evaluation of tested models.}
	\begin{center}
		\begin{tabularx}{\textwidth}{|X|X|X|X|X|X|X|p{2.4cm}|}
			\hline	
			Model & Training loss & Validation loss & Test loss & Training \newline accuracy & Validation \newline accuracy & Test accuracy & Overall practical \newline performance (1-5) \\ \hline
			Inception-ResNet & 0.201513 & 0.202228 & 0.211384 & 0.9990 & 0.9965 & 0.9700 & 5 \\ \hline
			ResNet152 & 0.202759 & 0.203262 & 0.214127 & 0.9985 & 0.9979 & 0.9711 & 4 \\ \hline
			Inception & 0.200859 & 0.210097 & 0.211211 & 1.0 & 0.9769 & 0.9711 & 3 \\ \hline
			MobileNet-Large & 0.280048 & 0.288908 & 0.288908 & 0.9451 & 0.9370 & 0.8872 & 2 \\ \hline
			MobileNet-Small & 0.311259 & 0.336001 & 0.332612 & 0.8224 & 0.6888 & 0.6752 & 1 \\ \hline
		\end{tabularx}
		\label{tab1}
	\end{center}
\end{table*}

\subsection{Graphical User Interface Description}

The GUI is designed using the Qt Creator Integrated Development Environment (IDE). Fig. \ref{fig:gui} displays the main screen of the GUI. An animated GIF file is played on the right side of the screen to guide the user through each step. The instruction of each stage is sequentially shown on the screen below the GIF file. The live stream of the camera video capture is shown on the left side of the screen, where the user can see her actions. A guideline is displayed at the bottom of the screen, helping users follow all stages. A progress indicator bar responds to the user activity when the step is performed correctly in a pre-specified time, and the current step is marked as passed. If there are unmarked steps, the user should repeat them at the end of the cycle.

\subsection{Deep Learning Model Description}

The deep neural network is a fundamental part of the system, which constantly classifies the user's hand rubbing steps and provides a corresponding output. This network is an Inception-ResNet-v2 \cite{b15} model, pre-trained on the ImageNet \cite{b16} database, and it is fine-tuned on our dataset and according to the application requirements. The input of the model is an RGB image, and the output is the probability of 9 classification categories (various hand rub steps). As in each step, the next is known, the Sigmoid activation function is used for the output layer. Moreover, a threshold on the output of the Sigmoid function should be set for considering the step passed. Therefore, different threshold values were tested to find and exploit the most effective one. 

Various related networks have been tested, and the results are illustrated in Table~\ref{tab1}. These models include two MobileNet \cite{b17} models (MobileNet-Small and MobileNet-Large), Inception \cite{b18}, ResNet152 \cite{b19}, and Inception-ResNet models. The practical performance metric measures the model generalization and is an assessment of the model's performance in a new environment which is relatively different from the dataset's environment, including different lighting, background, and image quality. This metric is empirical and is evaluated subjectively and manually. Volunteers tested the system based on each model individually and expressed their experience. Conclusions are made based on this data since determining an accurate metric is challenging. 

\begin{figure}[t]
	\centerline{\includegraphics[width=7.5cm]{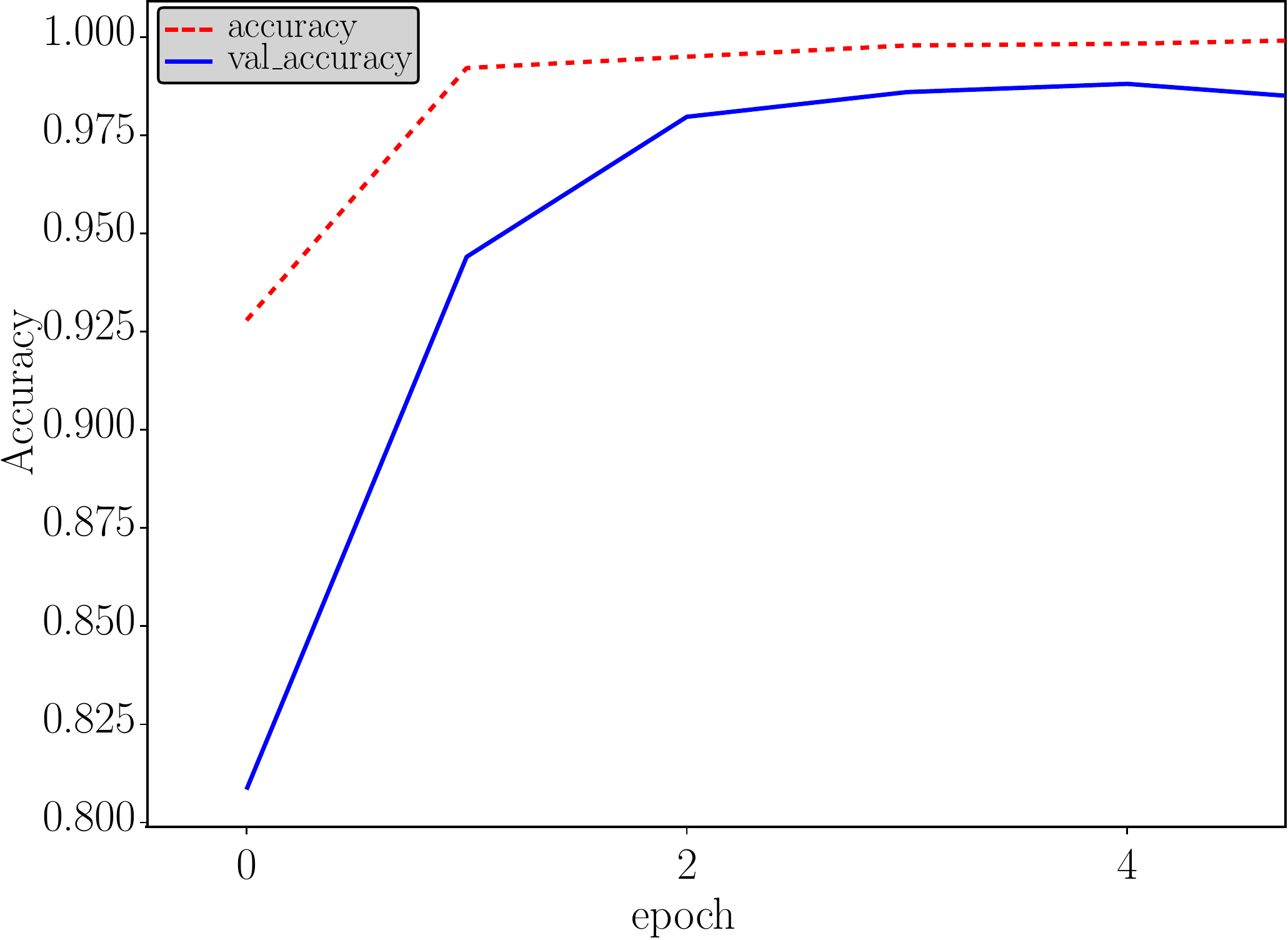}}
	\caption{The accuracy curve of the Inception-ResNet model.}
	\label{fig:accuracy}
\end{figure}

\begin{figure}[t]
	\centerline{\includegraphics[width=7.5cm]{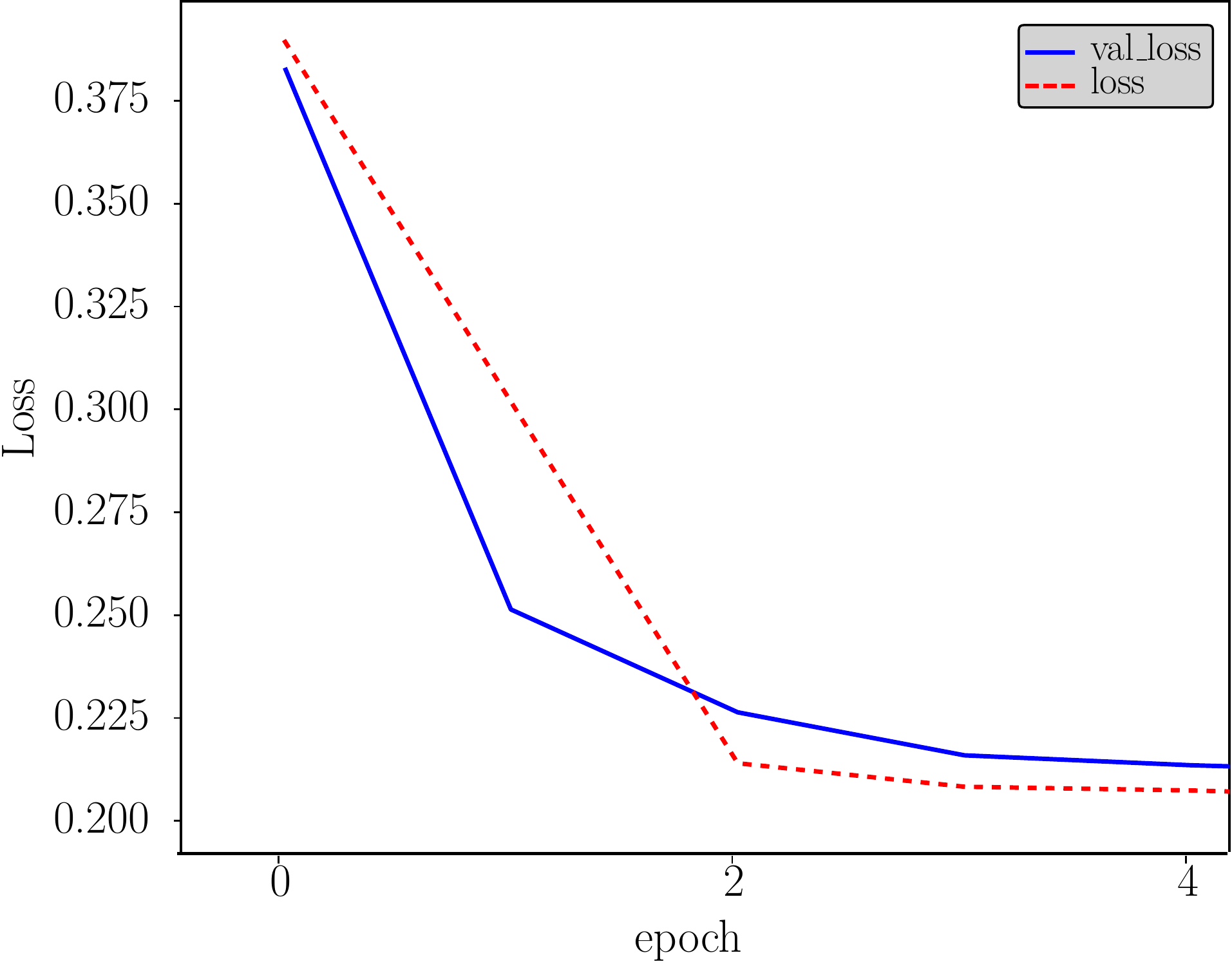}}
	\caption{The loss curve of the Inception-ResNet model.}
	\label{fig:loss}
\end{figure}

\begin{figure}[t]
	\centerline{\includegraphics[width=7.4cm]{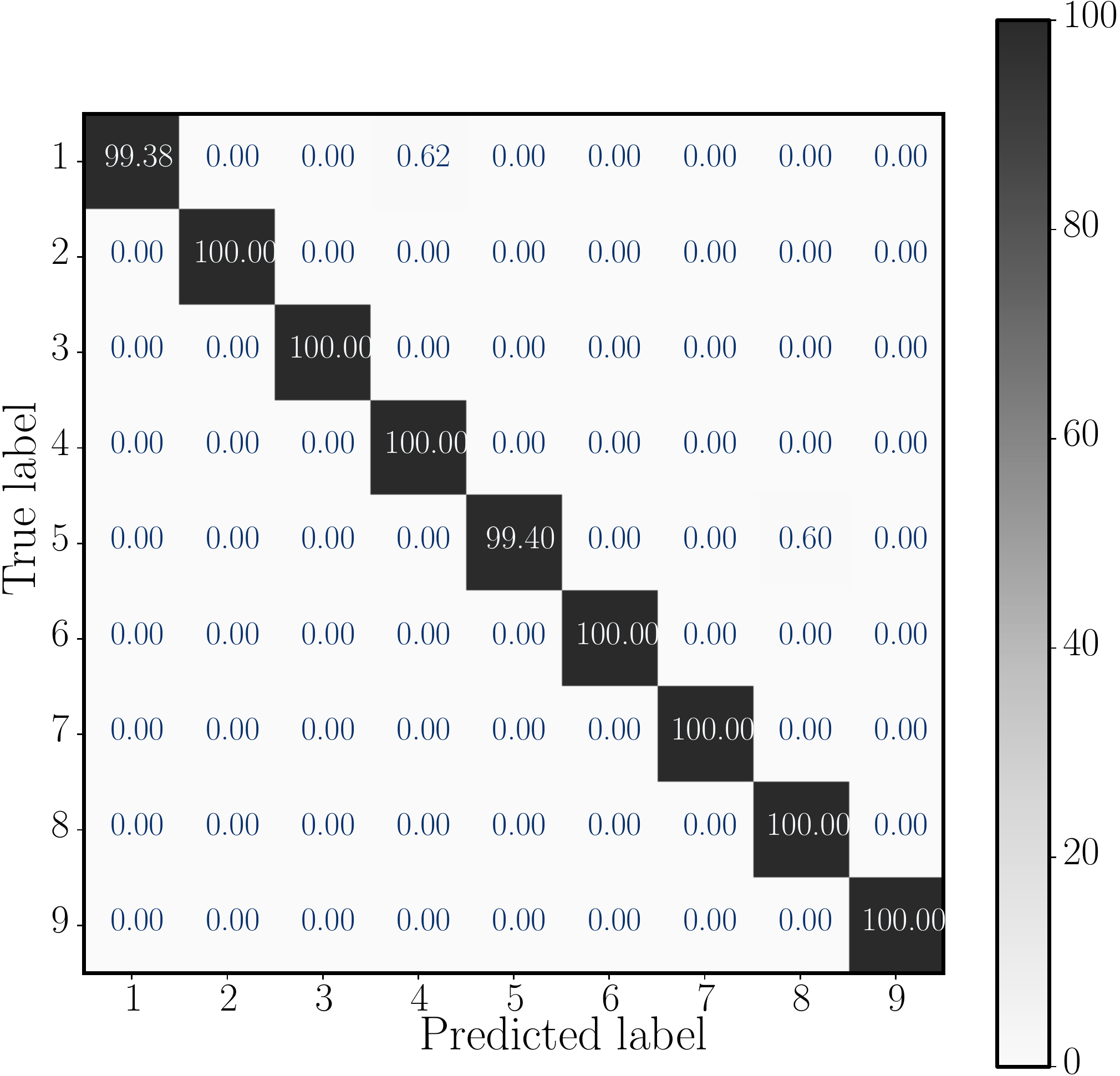}}
	\caption{The validation confusion matrix of the Inception-ResNet model.}
	\label{fig:confusionv}
\end{figure}

\begin{figure}[t]
	\centerline{\includegraphics[width=7.4cm]{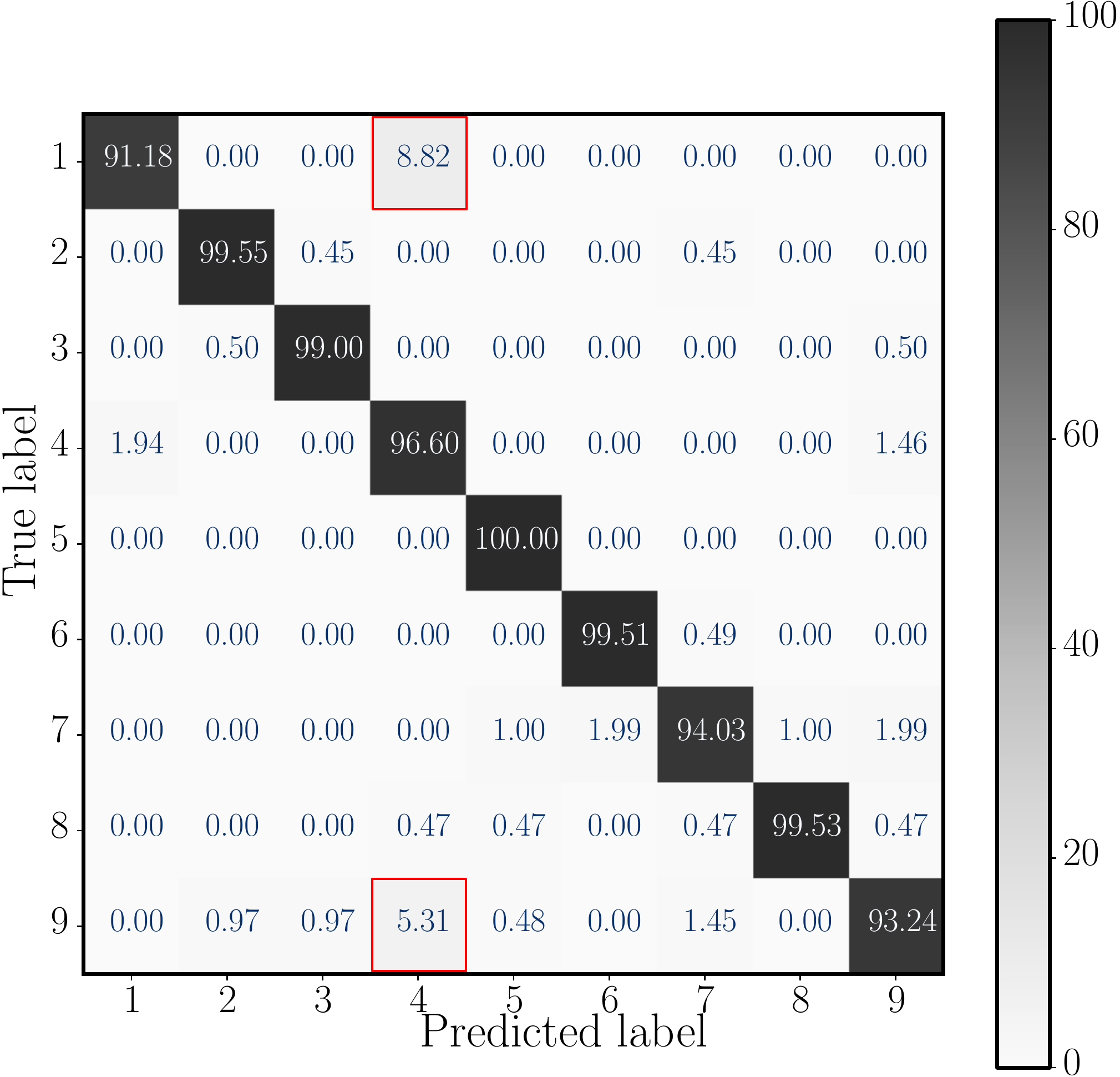}}
	\caption{The test confusion matrix of the Inception-ResNet model.}
	\label{fig:confusiont}
\end{figure}

As shown in Table~\ref{tab1}, the MobileNet models have lower accuracies compared to other models. This is justified because these networks are relatively smaller and designed for cases where the hardware is not high-performing. MobileNet-Large has higher accuracy than MobileNet-Small, but it is a deeper network and does more computation, thus slower. The Inception network has reached an accuracy of 100\% and seems to have overfitted. This network was tested in new environments and found out that it has a lower performance in comparison to ResNet and Inception-ResNet. The ResNet model performs well, but its loss is higher than Inception and Inception-ResNet models. It is also slow in training (its training time is almost 3.5 times of Inception and 1.5 times of Inception-ResNet). Finally, the Inception-ResNet model, which shares characteristics of both Inception and ResNet networks, has low loss and high accuracy (its accuracy is 0.11\% lower than Inception and ResNet, which might be a result of the randomness of the learning process). Also, its training time is fairly low, and it has a remarkably high performance in new environments. It is worthwhile to mention that in the actual setting of the built system, the background differs from the dataset, image quality is lower, and the lighting condition is different. Taking these differences into account, the Inception-ResNet model has a considerable generalization and correctly recognizes actions of users who were not present in the videos of the dataset. Additionally, this model's learning process is more stable, and its loss and accuracy curves do not include spikes and are far smoother than the ones of other models. Training and validation accuracy and loss values are close, indicating that the model has not overfitted the training data. Figs. \ref{fig:confusionv} and \ref{fig:confusiont} depict the validation and test confusion matrices, respectively. From the obtained results, it can be inferred that the proposed model can correctly recognize all of the steps. However, from the foregoing matrices, one can observe some confusion between steps 1 and 4, which is due to the physical similarity in performing these two steps. 

\subsection{Dataset Description}
\begin{figure}[t]
	\centerline{\includegraphics[width=6cm]{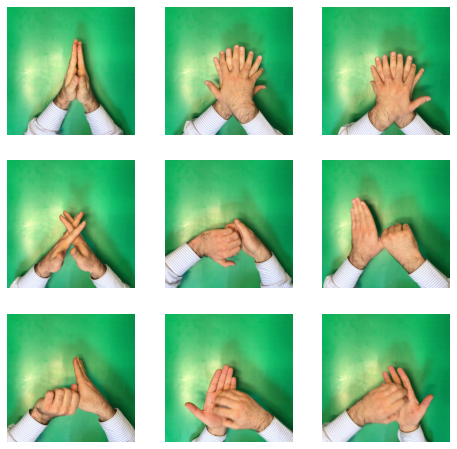}}
	\caption{Sample frames of dataset, green background.}
	\label{fig:datasetg}
\end{figure}

\begin{figure}[t]
	\centerline{\includegraphics[width=6cm]{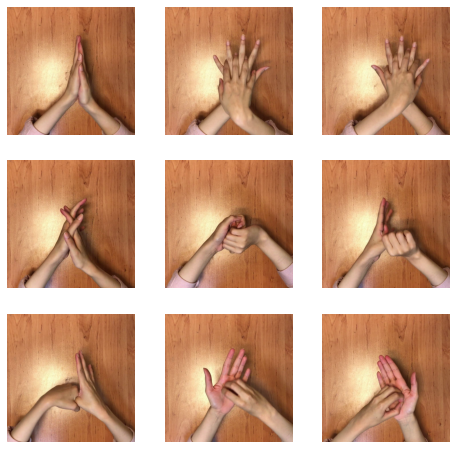}}
	\caption{Sample frames of dataset, wooden background.}
	\label{fig:datasetw}
\end{figure}

As a convenient dataset was required for this application and such a dataset was not available, videos of 22 volunteers, 18 men and 4 women, carrying out the hand rubbing steps as recommended by the guideline were recorded. For each volunteer, each step is recorded in two different environments. The first set of videos is recorded with wooden background, and the second one has a green background. Samples of the dataset are illustrated in Figs. \ref{fig:datasetg} and \ref{fig:datasetw}.

\section{Experimental Results}
\begin{figure}[tb]
	\centerline{\includegraphics[width=\linewidth]{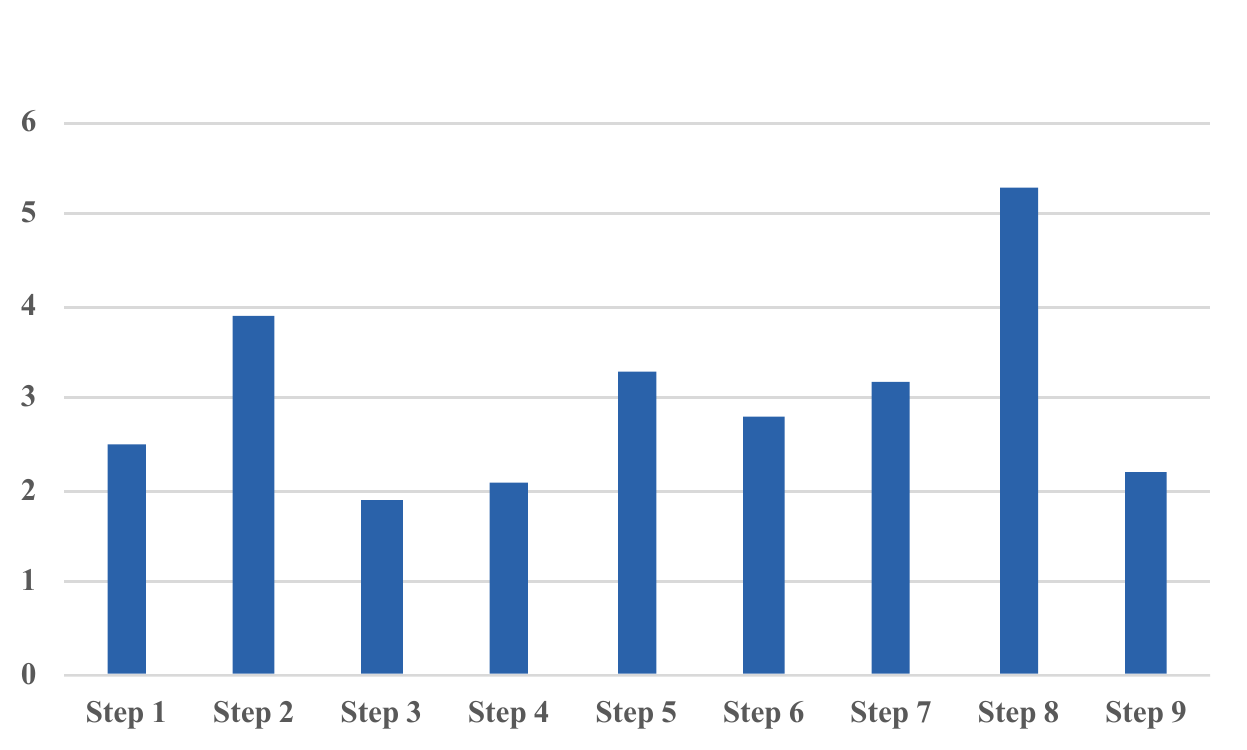}}
	\caption{Average time of each step in the experiment.}
	\label{fig:barchart}
\end{figure}

For ten participants, three females and seven males, a study is carried out to measure the average time of each step. Among these volunteers, five have used the system for the first time. Fig. \ref{fig:barchart} demonstrates the average required time for each hand rubbing step. It indicates that the most challenging step is step 8, taking 5.3 seconds on average. Other challenging steps are steps 2 and 5, taking 3.9 and 3.3 seconds on average, respectively. As suggested by WHO, the duration of the entire hand rubbing procedure should be between 20 and 30 seconds \cite{b6}. In this experiment, the average time of the hand rubbing procedure, excluding the sanitizer dispensing steps, is 27.2 seconds, which conforms to the standard timing of hand washing in the WHO guidelines.

\section{Conclusion}
In conclusion, an automated hand hygiene training system was proposed in this paper. The approach was based on markerless hand gesture recognition using an Inception-ResNet DNN, which has yielded the best results among evaluated models. In the system, user hand presence was first detected, and then alcohol was dispensed. Afterward, hand rubbing steps were shown on the screen, and as the user followed the steps correctly, the GUI was updated. The system was fabricated, and the main computing unit of the system was an NVIDIA Jetson AGX Xavier. The system was evaluated in a real-life scenario of being used by volunteers. The average of the overall time taken by the hand rubbing steps in this empirical testing was 27.2 seconds, which is within the interval suggested by WHO. Furthermore, a dataset was collected, including videos of volunteers carrying out the hand rubbing steps as suggested by the WHO guidelines. Future work includes the detection of fake gestures since an expert user might try to trick the system by making similar but incorrect gestures. In addition, an embedded system with lower performance than the NVIDIA Jetson AGX Xavier, e.g., the NVIDIA Jetson Nano, might be used instead if further optimizations are carried out.

\end{document}